\documentclass[sigconf]{acmart}

\usepackage{acmart-taps}
\usepackage{stfloats}

\usepackage{microtype}
\usepackage{balance}
\usepackage{booktabs}
\usepackage{graphicx}
\usepackage{multirow}

\AtBeginDocument{%
  \providecommand\BibTeX{{%
    \normalfont B\kern-0.5em{\scshape i\kern-0.25em b}\kern-0.8em\TeX}}}

\makeatletter
\def\@ACM@copyright@check@cc{}
\makeatother

\copyrightyear{2025}
\acmYear{2025}
\setcopyright{cc}
\setcctype{by}
\acmConference[CHI '25]{CHI Conference on Human Factors in Computing Systems}{April 26-May 1, 2025}{Yokohama, Japan}
\acmBooktitle{CHI Conference on Human Factors in Computing Systems (CHI '25), April 26-May 1, 2025, Yokohama, Japan}\acmDOI{10.1145/3706598.3713330}
\acmISBN{979-8-4007-1394-1/2025/04}

\acmSubmissionID{7119}

\begin{document}

\title[\texttt{SET-PAiREd}: Designing for Parental Involvement with an AI-Assisted Educational Robot]{\texttt{SET-PAiREd}: Designing for Parental Involvement in Learning with an AI-Assisted Educational Robot}

\author{Hui-Ru Ho}
\orcid{0009-0000-3701-2521}
\affiliation{%
  \institution{Department of Computer Sciences\\
  University of Wisconsin--Madison}
  \city{Madison}
  \state{Wisconsin}
  \country{USA}
}
\email{hho24@cs.wisc.edu}

\author{Nitigya Kargeti}
\orcid{0000-0001-5970-7332}
\affiliation{%
  \institution{Department of Computer Sciences\\
  University of Wisconsin--Madison}
  \city{Madison}
  \state{Wisconsin}
  \country{USA}
}
\email{kargeti@wisc.edu}

\author{Ziqi Liu}
\orcid{0009-0007-8755-5744}
\affiliation{%
  \institution{Department of Computer Sciences\\
  University of Wisconsin--Madison}
  \city{Madison}
  \state{Wisconsin}
  \country{USA}
}
\email{ziqil@cs.wisc.edu}

\author{Bilge Mutlu}
\orcid{0000-0002-9456-1495}
\affiliation{%
  \institution{Department of Computer Sciences\\
  University of Wisconsin--Madison}
  \city{Madison}
  \state{Wisconsin}
  \country{USA}
}
\email{bilge@cs.wisc.edu}


\renewcommand{\shortauthors}{Ho et al.}

\begin{abstract} 
AI-assisted learning companion robots are increasingly used in early education. Many parents express concerns about content appropriateness, while they also value how AI and robots could supplement their limited skill, time, and energy to support their children's learning. We designed a card-based kit, \texttt{SET}, to systematically capture scenarios that have different extents of parental involvement. We developed a prototype interface, \texttt{PAiREd}, with a learning companion robot to deliver LLM-generated educational content that can be reviewed and revised by parents. Parents can flexibly adjust their involvement in the activity by determining what they want the robot to help with. We conducted an in-home field study involving 20 families with children aged 3--5. Our work contributes to an empirical understanding of the level of support parents with different expectations may need from AI and robots and a prototype that demonstrates an innovative interaction paradigm for flexibly including parents in supporting their children.
\end{abstract}



\begin{teaserfigure}
  \includegraphics[width=\textwidth]{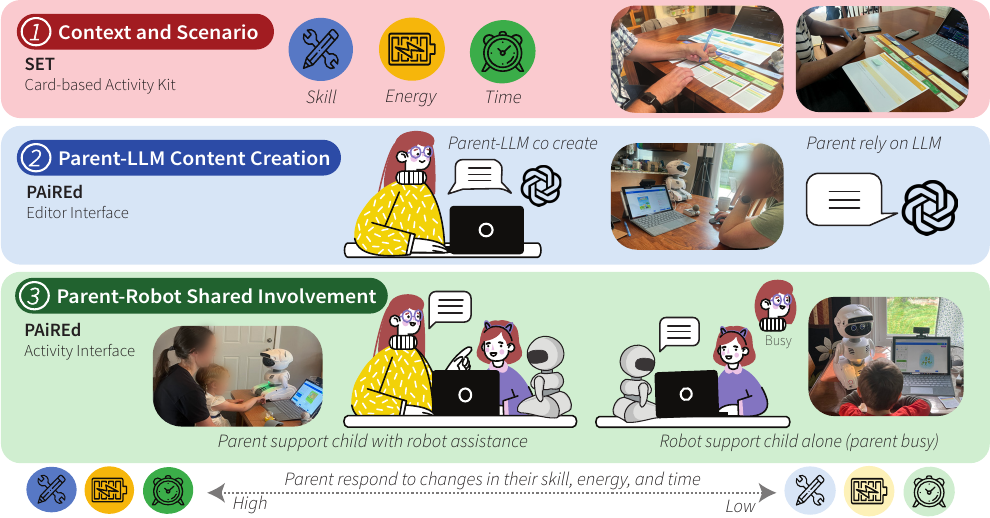}
  \caption{We explored parental involvement scenarios using the \texttt{SET} activity kit, examined parents' perceptions of AI-generated content, and analyzed their use of \texttt{PAiREd} for collaborating with LLM and robots to support their children's learning.}
  \label{fig:individual}
\end{teaserfigure}
\sloppy
\maketitle

\section{Introduction}

Parental involvement is essential to children's early education \cite{ma2016meta, harris2008parents}. However, despite parents' desire to participate in their children's learning, many of them face challenges, including limited capability, availability, and motivation \cite{green2007parents, ho2024s, hara1998parent}, which hinder effective participation. For example, one parent may struggle to come up with developmentally appropriate questions to ask their child, while another may know what to discuss but feel unable to dedicate sufficient time due to demanding work-home schedules. 

AI-assisted robots hold significant potential to transform both the scope and methods of parental involvement in children's early education. Prior work has highlighted that AI-powered systems can personalize educational experiences to address individual learning needs \cite{zhang2024mathemyths, xu2022elinor, xu2024artificial} and that social robots can facilitate rich interactions that support children's cognitive and emotional development in the physical world \cite{ho2023designing, leyzberg2012physical, belpaeme2018social, kim2024understanding, lee2022unboxing}. Yet, existing studies have primarily focused on addressing child-centric needs, overlooking parental roles. Additionally, parents often raise concerns about privacy, misinformation, and content appropriateness \cite{oswald2020psychological, david2007electronic, howard2021digital}, emphasizing the need for parental supervision to ensure alignment of AI and robots with parents' educational goals \cite{ho2024s, han2024teachers}.

In light of these gaps, it is essential to understand the contexts in which parents support their children's learning, investigate how AI-assisted robots can be designed to help parents overcome barriers in these contexts, and address concerns regarding children's interactions with these technologies. We propose that providing parents with \textit{flexible control} over AI-generated content and \textit{collaborative involvement} with educational robots can effectively support parents in responding to dynamic contexts and barriers, balancing the benefits and risks of AI-assisted robots. More specifically, this paper explores the following research questions:
\begin{itemize}
    \itemsep0.4em
    \item \textbf{RQ1:} What contexts do parents encounter when involving in young children's learning activities?
    \item \textbf{RQ2:} How do parents perceive AI-generated learning content for young children?
    \item \textbf{RQ3:} How would parents prefer to collaborate with LLM on supervising content creation under different contexts?
\end{itemize}

\begin{itemize}
    \item \textbf{RQ4:} How would parents prefer to collaborate with an AI-assisted robot to engage in learning activities with their children under different contexts?
\end{itemize}

To address these questions, we developed two tools: (1) a card-based activity kit named \texttt{SET} for systematically understanding parental involvement contexts, and (2) a prototype AI-assisted robot named \texttt{PAiREd} to explore flexible parent-AI and parent-robot collaboration mechanisms. The \texttt{SET} kit, with its name representing the key factors of \textbf{S}kill, \textbf{E}nergy, and \textbf{T}ime \cite{ho2024s, green2007parents}, helps parents reflect on their experiences in supporting their children's learning. Parents use the kit to create scenario cards that capture relatable contexts framed around these factors, serving as both study findings and research tools for contextualizing parent's interaction with the AI-assisted robot. In addition, the \texttt{PAiREd} prototype (\textbf{P}arenting with \textbf{Ai}-assisted \textbf{R}obot for \textbf{Ed}ucation) enables parents to generate, edit, and revise storybook content in collaboration with a large language model (LLM). During parent-child reading sessions, parents can adjust task delegation between themselves and the robot via a flexible mechanism, allowing control over the autonomy of the AI-assisted robot. This approach allows parents to balance their involvement based on the specific needs of different scenarios.

We evaluated the \texttt{SET-PAiREd} system through a four-stage study. First (\textbf{RQ1}), parents used \texttt{SET} to document real-life involvement scenarios, revealing contextual needs and scenarios. Second (\textbf{RQ2}), interviews about AI-generated content showed that parents value adaptability, efficiency, and affordability, but worry about age appropriateness, inaccuracies, and over-reliance. Third (\textbf{RQ3}), parents co-created learning content for two books using \texttt{PAiREd}, indicating that difficulty, variety, and quality drive their evaluation of LLM-generated content. Under the \texttt{SET} scenarios, their collaboration with the LLM varied depending on their time, energy, trust in the LLM, and interest in editing content. Finally (\textbf{RQ4}), parents read the books with their children, adjusted role delegation between themselves and the robot according to the scenarios, and provided feedback. We identified several collaboration patterns, including using \textit{parent takeover} mode with LLM-generated content, \textit{collaborative} modes guided by parent skill and motivation, and \textit{robot takeover} mode under parental supervision or with reviewed LLM content. Our work makes the following contributions:
\begin{enumerate} 
    \item \textit{Prototype Artifact}: We developed the \texttt{SET} activity kit to map parents' real-life contexts around their skill, energy, and time. In addition, we developed and tested \texttt{PAiREd}, a system supporting responsible supervision of LLM-generated content and enabling flexible parent-robot collaboration.
    \item \textit{Novel Insights}: Field testing provided insights into systematic understanding of parental involvement scenarios through \texttt{SET} (RQ1), parents' perceptions of AI-generated content (RQ2), parents' preferences for \texttt{PAiREd} during parent-AI collaboration in content creation (RQ3), and parent-robot collaboration in facilitating learning activities (RQ4). The insights reveal novel interaction paradigms for human-robot collaboration in the context of parenting and education.
    \item \textit{Design Recommendations}: We synthesized our findings into actionable recommendations to enhance parental involvement through AI-assisted learning companion robots.
\end{enumerate}
\section{Related Work}

In this section, we review research related to the importance and barriers to parental involvement; parental use of learning technologies; and the use of generative AI and robot in educational and parenting scenarios.

\subsection{Importance and Barriers to Parental Involvement}\label{sec-rw-2.1}

Early childhood is a critical period for predicting future success and well-being, with early education investments resulting in higher returns than later interventions \cite{duncan2007school, doyle2009investing}. Effective parental involvement fosters cognitive and social skills, especially in younger children \cite{blevins2016early, peck1992parent}. Parents are encouraged to prioritize home-based involvement to maximize their influence \cite{ma2016meta}, as their involvement has a greater impact on children's learning outcomes \cite{hoffner2002parents, fehrmann1987home, hill2004parent} within the family setting than partnerships with schools or communities \cite{ma2016meta, harris2008parents, fantuzzo2004multiple, sui1996effects}.

However, parents' involvement in their children's education is often constrained by practical challenges related to parents' \textit{skills}, \textit{time}, and \textit{energy}. The Hoover-Dempsey and Sandler (HDS) framework \cite{green2007parents} and the CAM framework \cite{ho2024s} both highlight these factors-- parents' perceived \textit{skills and knowledge} (capability), \textit{time} (availability), and \textit{motivation} (energy)--influence the extent of their engagement. For instance, a parent confident in math may choose to engage more in math-related tasks, while those facing inflexible schedules may participate less \cite{green2007parents}. Unlike teachers, parents often lack formal pedagogical training and may underestimate their role in supporting children's learning, particularly as young children struggle to articulate their needs \cite{hara1998parent}. The CAM framework similarly suggests parents delegate tasks to a robot when they feel less capable, have limited time, or are unmotivated. These factors reflect parents' life contexts, shaped by demographic backgrounds, occupations, and parenting responsibilities \cite{grolnick1997predictors}, highlighting the need to help parents overcome barriers to effective involvement in early education within their life contexts.

\subsection{Parental Use of Learning Technologies}\label{sec-rw-2.2}
Technology encourages parental involvement by facilitating parent-child engagement in learning activities while introducing risks that require active parental mediation \cite{gonzalez2022parental}. On the positive side, technology offers novel opportunities for parental engagement and enhances children's learning outcomes. For example, e-books promote interactive behaviors between parents and children better than print books \cite{korat2010new}. In addition, having access to computers at home significantly boost academic achievement of young children when parents actively mediate their use \cite{hofferth2010home, espinosa2006technology}. However, the effectiveness of these tools often depends on parents' familiarity with and attitude toward technology. Mobile applications, for instance, can improve learning outcomes but require parents to possess sufficient technology efficacy to guide their use \cite{papadakis2019parental}.

On the negative side, technology introduces risks such as excessive screen time, exposure to inappropriate content, and misinformation, which necessitate parental intervention \cite{oswald2020psychological, howard2021digital}. According to parental mediation theory, parents mitigate these risks through restrictive mediation (e.g., setting limits), active mediation (e.g., discussing content), and co-use (e.g., shared use of technology) \cite{valkenburg1999developing}. Modern technologies like video games, location-based games (\textit{e.g.,} Pokemon Go), and conversational agents (\textit{e.g.,} Alexa) also require parents to adapt their mediation strategies to ensure responsible use \cite{valkenburg1999developing, nikken2006parental, sobel2017wasn, beneteau2020parenting, yu2024parent}. Overall, parents seek to leverage technology to support their children's learning due to ite effectivenss but are also mindful of its risks. Their involvement is therefore driven by both opportunities and concerns, highlighting the need to design tools that effectively involve parents to balance benefits and risks.

\subsection{Generative AI and Companion Robots for Parenting and Education}
Generative AI and companion robots offer human-like affordances, with AI simulating human intelligence and robots providing physical human-like features. Compared to conventional models (\textit{e.g.,} machine learning) and devices (\textit{e.g.,} laptops), these emerging technologies enable natural and social interactions, creating opportunities for novel paradigms to enhance parental involvement and children's learning while introducing their unique challenges.

\subsubsection{Generative AI}
GAI offers promising support for parents by enhancing their ability to educate and engage with their children. Prior work suggested that AI-driven systems can support parenting education \cite{petsolari2024socio} and provide evidence-based advice through applications and chatbots, delivering micro-interventions such as teaching parents how to praise their children effectively \cite{davis2017parent, entenberg2023user} or offering strategies to teach complex concepts \cite{mogavi2024chatgpt, su2023unlocking}. Many parents also prefer using GAI to create educational materials tailored to their children's needs, rather than granting children direct access to these tools \cite{han2023design}. Beyond educational support, AI-based storytelling tools address practical challenges (\textit{e.g.,} time constraints) by alleviating physical labor while fostering parent-child interactions \cite{sun2024exploring}. Furthermore, GAI offers advantages to children's learning directly. It can help create personalized learning experiences by providing timely feedback and tailoring content \cite{su2023unlocking, mogavi2024chatgpt, han2024teachers}, enhancing positive learning experiences \cite{jauhiainen2023generative}. For example, a LLM-driven conversational system can teach children mathematical concepts through co-creative storytelling, achieving learning outcomes similar to human-led instructions \cite{zhang2024mathemyths}.

Despite these benefits, several concerns persist regarding the use of GAI in education. Prior work highlighted the limitations of GAI, such as its limited effectiveness in more complex learning tasks,the limited quality of the training data, and its inability to offer comprehensive educational support \cite{su2023unlocking}. There is also a significant risk of GAI producing inaccurate or biased information, discouraging independent thought among children, and threatening user privacy \cite{su2023unlocking, han2023design, han2024teachers}. Many parents are skeptical about the role of AI in their children's academic processes, concerned about the accuracy of AI-generated content, and worry that over-reliance on AI could stifle independent thinking \cite{han2023design}.

\subsubsection{Social companion robots}
Social companion robots have proven potential to assist parents in home education settings through studies in \textit{parent-child-robot} interactions. \citet{gvirsman2020patricc} showed that the robotic system, \texttt{Patricc}, fostered more triadic interaction between parents and toddlers than a tablet, and \citet{gvirsman2024effect} found that, in a parent-toddler-robot interaction, parents tend to decrease their scaffolding affectively when the robot increases its scaffolding behavior. Similarly, \citet{chen2022designing} found that social robots enhanced parent-child co-reading activities, while \citet{chan2017wakey} demonstrated that the WAKEY robot improved morning routines and reduced parental frustration. Beyond educational support, \citet{ho2024s} uncovered that parents envisioned robots as their \textit{collaborators} to support their children's learning at home and that their collaboration patterns can be determined by the parents' capability, availability, and motivation. Although parents generally have positive attitudes toward incorporating robots into their children's learning, they remain concerned about the risk of disrupting school-based learning and potential teacher replacement \cite{tolksdorf2020parents, lee2008elementary, louie2021desire}.

In addition to parental support, social companion robots also support children in education directly through \textit{child-robot interactions}. Physically embodied robots provide adaptive assistance and verbal interaction similar to virtual or conversational agents \cite{ramachandran2019personalized, leyzberg2014personalizing, schodde2017adaptive, brown2014positive}, yet they foster greater engagement with the physical environment and encourage more advanced social behaviors during learning \cite{belpaeme2018social}, leading to improved learning outcomes \cite{leyzberg2012physical}. Prior work demonstrated that companion robots can effectively support both school-based learning (\textit{e.g.,} math \cite{lopez2018robotic}, literacy \cite{kennedy2016social, gordon2016affective}, and science \cite{davison2020working}) and home-based learning activities (\textit{e.g.,} reading \cite{michaelis2018reading, michaelis2019supporting}, number board games \cite{ho2021robomath}, and math-oriented conversations with parents \cite{ho2023designing}). For example, \citet{kennedy2016social} suggested that children can learn elements of a second language from a robot in short-term interactions, and \citet{tanaka2009use} found that children who took on the role of teaching the robot gained confidence and improved learning outcomes.


Parental involvement in early education is crucial and AI-assisted robots can offer promising support by helping parents overcome practical barriers (\textit{i.e.,} time, energy, and skills) and addressing concerns about technological risks. Yet, limited research has examined how technology design can simultaneously alleviate these barriers and concerns. Though \citet{zhang2022storybuddy} emphasized the importance of flexible parental involvement during reading through a system called \textit{Storybuddy}, yet they focused on a virtual chatbot rather than a physical robot, and how the flexible modes may be used in different scenarios remain unknown. Similarly, \textit{ContextQ} \cite{dietz2024contextq} presented auto-generated dialogic questions to caregivers for dialogic reading, but primarily considered situations where parents are actively involved, not scenarios where parents cannot participate fully.

In this work, we address these gaps by exploring parental involvement contexts, understanding parents' perceptions of AI-generated content, and examining how parents collaborate with AI and robots under different scenarios. In the following sections, we describe our development of  \texttt{SET}, a card-based activity, to understand parental involvement contexts (Section~\ref{sec-card}), the design of the \texttt{PAiREd} system to enable parents to co-create learning activities with an LLM (Section~\ref{sec-system}), and user study aimed to discover use patterns and understand user perceptions of the system (Section~\ref{sec-study}).
\section{Card-Based Activity Kit Design} \label{sec-card}

\begin{table*}[!t]
    \caption{Factors and dimensions used in the Card-based Activity Kit}
    \label{tab:kit-example}
    \centering
    \small
    \begin{tabular}{p{0.12\linewidth}p{0.4\linewidth}p{0.4\linewidth}}%
        \toprule
         \textbf{Factor-level} & \textbf{Description} & \textbf{Example} \\
        \midrule
         Skill-high & Parents feels \textit{very} confident in a specific task or topic. & ``\textit{I am good at math-related activities.}'' \\
         \midrule
         Skill-low & Parents feels \textit{not} confident in a specific task or topic. & ``\textit{I am not good at coming up with good questions to ask.}'' \\
         \midrule
         Energy-high & Parents feels \textit{highly} motivated. & ``\textit{I am highly motivated when my child invites me.}'' \\
         \midrule
         Energy-low & Parents feels \textit{not} motivated. & ``\textit{I am not motivated when I am tired.}'' \\
         \midrule
         Time-high & Parents are \textit{available and present}. & ``\textit{I am available and present when my work is done.}'' \\
         \midrule
         Time-low & Parents \textit{need to be absent}. & ``\textit{I need to be absent when my younger child is crying.}'' \\     
         \bottomrule
    \end{tabular}
\end{table*}

To systematically explore parents' contextual, real-life scenarios where varying levels of parental involvement in children's learning may occur, we designed a card-based activity kit named \textbf{\texttt{SET}}, representing \textbf{\texttt{S}}kill, \textbf{\texttt{E}}nergy, and \textbf{\texttt{T}}ime. The design of the kit is grounded in frameworks from education \cite{green2007parents} and HCI \cite{ho2024s}. Specifically, we leveraged the ``\textit{Parent Perceived Life Context}'' construct from the Hoover-Dempsey and Sandler (HDS) framework for parental involvement \cite{green2007parents} and the key factors influencing parent-robot collaboration defined by \citet{ho2024s}. We detailed the theoretical foundation in Section \ref{sec-rw-2.1}.

The kit consists of three primary materials: \texttt{SET-banners}, \texttt{SET-} \texttt{experience banks}, and \texttt{SET-scenario cards}\footnote{\url{https://osf.io/zfksg/?view_only=b59bd41287f543ce82ab85950aaf004f}}. Each material is characterized by the same two-dimensional factors (\textit{i.e.,} skill (high/low), energy (high/low), and time (high/low)) but is designed in different formats to serve distinct purposes during the card activity (see Table \ref{tab:kit-example} for details about each factor and its dimensions). Although the factors theoretically exist on a continuous spectrum, we divided them into binary dimensions to ensure the scale of the activity is manageable for participants. Below, we describe the purpose of each material. For more details on how the kit was used, refer to Section \ref{sec-procedure}.

\begin{figure*}[b!]
  \includegraphics[width=\textwidth]{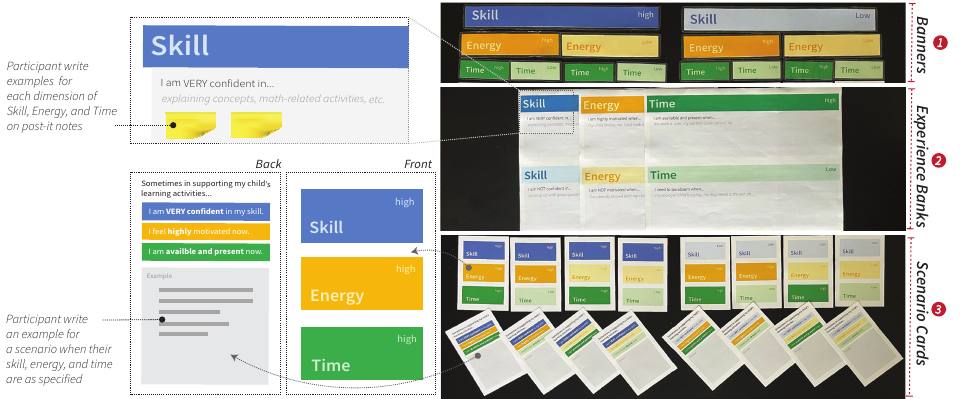}
   \vspace{-6pt}
  \caption{The \texttt{SET} Card-Based Activity Kit. (1) Banners are used to visualize a hierarchical diagram. (2) Experience banks allow participants to generate examples for each factor dimension. (3) Scenario cards guide participants to create real-life scenarios.}
  \label{fig:card-kit}
  \vspace{-3pt}
\end{figure*}

\begin{enumerate}
    \item The \texttt{SET-banners} present a hierarchical diagram that depicts how the three two-dimensional factors combine to create \textit{eight} distinct scenarios (Figure \ref{fig:card-kit}), forming the \textit{eight} \texttt{SET-scenario cards}.
    \item The \texttt{SET-experience banks} enable participants to document real-life examples for each dimension of every factor using post-it notes. This activity encourages focused reflection on individual factors and dimensions, serving as a stepping stone for constructing scenarios that integrate all three factors during the \texttt{SET-scenario cards} activity (Figure \ref{fig:card-kit}).
    \item The \texttt{SET-scenario cards} guide participants in creating \textit{eight} real-life scenarios, each defined by unique combinations of the three two-dimensional factors. The front of each card displays the dimensional status of the factors, while the back includes detailed descriptions of each factor's status and a blank space for participants to document a scenario that aligns with the corresponding dimensions (Figure \ref{fig:card-kit}).
\end{enumerate}

\section{System Design} \label{sec-system}

\begin{figure*}[t!]
  \includegraphics[width=\textwidth]{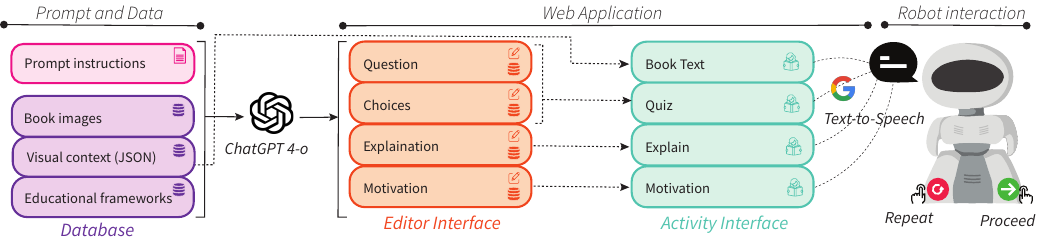}
   \vspace{-6pt}
  \caption{\texttt{PAiREd} System Architecture Overview. The system consists of three key components: prompt instructions and data, the web application (which includes both the editor and activity interfaces), and the robot interaction module.}
  \label{fig:syste-diagram}
   \vspace{-10pt}
\end{figure*}

We designed a prototype system, \texttt{PAiREd}, an LLM-driven interface integrated with an educational robot that uses LLMs to generate learning content based on established educational frameworks for reading activities. The system enables parents to review and revise LLM-generated learning content \textit{before} the activity and adjust their involvement by modifying the role delegation between the robot and themselves \textit{during} the activity. The system consists of three core design components: (1) \textit{LLM-driven content generation}, (2) \textit{Interface design}, and (3) \textit{Robot interaction design} (Figure \ref{fig:syste-diagram}).

\subsection{LLM-driven Content Generation}
Content generation is powered by a state-of-the-art LLM, ChatGPT-4o \cite{achiam2023gpt} developed by OpenAI\footnote{OpenAI ChatGPT-4o: \url{https://openai.com/index/hello-gpt-4o/}} through prompt engineering \cite{sahoo2024systematic}. To generate the learning content for each book page, the input we provided for GPT-4o includes: (1) \textit{prompts}\footnote{Prompts: \url{https://osf.io/zfksg/?view_only=b59bd41287f543ce82ab85950aaf004f}}: instructions to generate learning content, (2) \textit{book image}: an image of the chosen book page, (3) \textit{visual context}\footnote{Visual context: \url{https://osf.io/zfksg/?view_only=b59bd41287f543ce82ab85950aaf004f}}: a manually-created JSON-based dataset containing structured visual information of the book page, and (4) \textit{educational frameworks} \footnote{Frameworks: \url{https://osf.io/zfksg/?view_only=b59bd41287f543ce82ab85950aaf004f}}: theoretical-driven frameworks defining critical pre-school concepts for math and literacy. The books, visual contexts, and educational frameworks are placed in our database and retrieved using prompt instructions. We explain \textit{visual context} and \textit{educational frameworks} below.

\subsubsection{Visual Context} 
The ChatGPT-4o model faces limitations in quantitative and spatial reasoning on images and recognition of animated or anthropomorphic drawings (\textit{e.g.,} mistaking clothed animals for humans), limiting its ability to accurately auto-generate question-answer pairs for literacy and math based on storybook images. To enhance content generation accuracy, we created a JSON-based dataset to capture the \textit{visual context} of each storybook page. Each page is represented as a unique entity with detailed descriptions of objects (\textit{e.g.,} characters, animals, and environmental elements), properties, as well as their spatial relationships. This context was embedded in the prompt design, instructing GPT-4o to retrieve precise object properties such as color, count, and location. Integrating this structured approach significantly improved the accuracy of ChatGPT-4o to create question-answer pairs based on storybooks, reducing object-identification errors and enhancing consistency across pages. Systematic evaluations by the first and third authors confirmed these improvements.


\begin{figure*}[b!]
  \includegraphics[width=\textwidth]{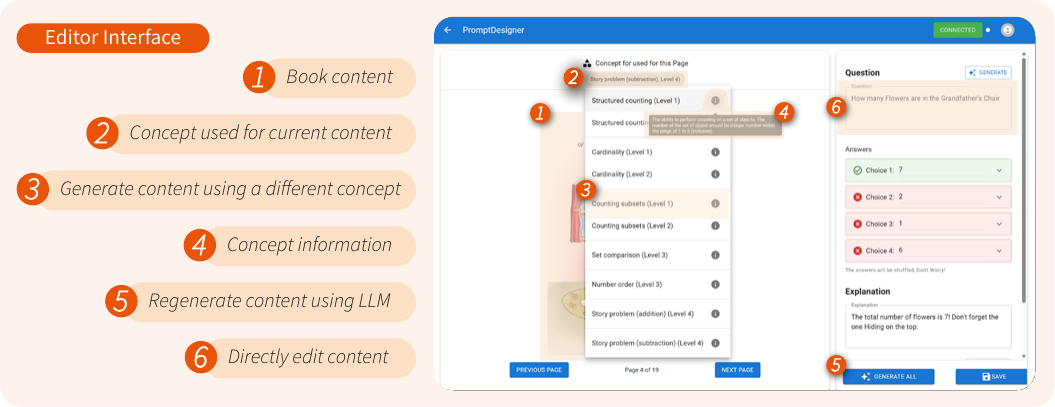}
     \caption{Editor Interface of the PAiREd system. This interface allows users to navigate book content, modify LLM-generated learning content through regeneration, or manually edit it.}
  \label{fig:editor}
\end{figure*}

\begin{figure*}[b!]
\centering
  \includegraphics[width=\textwidth]{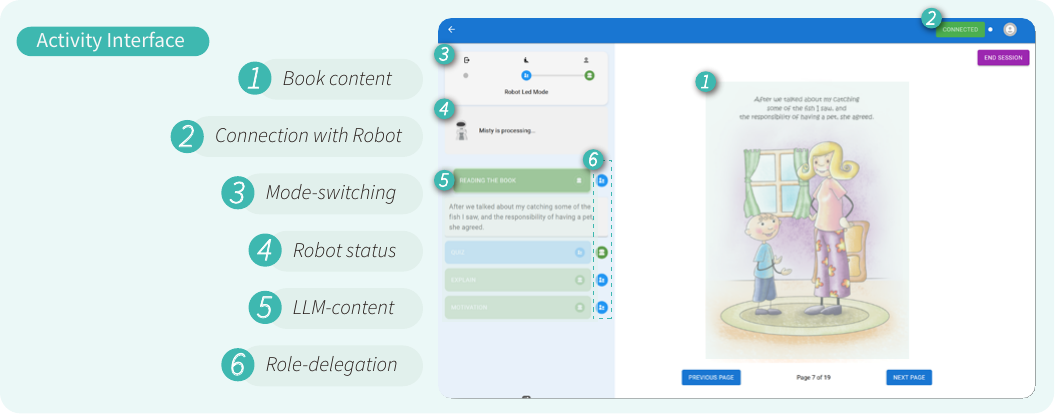}
  \caption{Activity Interface of the PAiREd system. This interface enables parents to flexibly adjust their involvement and seamlessly share responsibilities with the robot through the mode-switching and role-delegation mechanisms.}
  \label{fig:activity}
   \vspace{-10pt}
\end{figure*}
 
\subsubsection{Educational Frameworks for Math and Literacy} 
We developed frameworks outlining varying proficiency levels in math \cite{purpura2013informal, engel2013teaching} and literacy \cite{kaminski2014preschool} concepts for preschoolers, drawing on educational psychology research. Math and literacy were selected as they are foundational skills for preschoolers \cite{weiland2013impacts}. The math framework incorporated informal numeracy skills identified by \citet{purpura2013informal} and aligned them with the four proficiency levels proposed by \citet{engel2013teaching}. The literacy framework was based on the PELI assessment benchmarks \cite{kaminski2014preschool}.

\subsection{Interface Design}
We developed a web-based interface using a React front-end\footnote{\url{https://reactnative.dev}} and a FAST API back-end,\footnote{\url{https://fastapi.tiangolo.com}} supported by a MongoDB database.\footnote{\url{https://www.mongodb.com}} The interface includes three key components: (1) \textit{LLM content generation}: parents can request LLM to generate a new set of learning content (\textit{e.g.,} question-answer pairs); (2) \textit{Editor interface}: parents can review and revise LLM-generated content to varying extents; and (3) \textit{Activity interface}: parents can flexibly adjust their involvement and role delegation between themselves and the robot. 

\subsubsection{LLM Content Generation}
Parents can choose a book from the digital library in the system and generate a new set of learning content using the LLM embedded within the platform. Before generating content, parents are prompted to select \textit{grade level} (\textit{e.g.,} preschool) and \textit{subject} (\textit{e.g.,} math or literacy). Once the content is generated, parents can either \textit{edit and review} the material or \textit{launch} it directly. For each page of the book, the LLM creates content following a structured approach, including three main components: (1) \textit{question and multiple choices}: a question related to a selected concept from the educational framework, with four answer options, one being correct; (2) \textit{explanation}: a description of how the correct answer is derived; (3) \textit{motivation}: encouraging words or a fun fact related to the content of the book page.

\subsubsection{Editor Interface}
After the LLM generated the content, if the user choose to \textit{edit and review}, the editor interface will display the content page by page. The main \textit{book content} is shown in the left panel, while the \textit{LLM-generated content} appears on the right. The \textit{concept}, initially selected by the LLM from the chosen educational framework, is displayed at the top of the book content. For the \textit{book content}, parents could navigate through the book and review it page by page. The \textit{LLM-generated content} was organized into its main components: \textit{questions and choices}, \textit{explanations}, and \textit{motivational content}. Parents had the option to either regenerate individual components or all components at once using the LLM, or manually edit the content. Additionally, the \textit{concept} was presented as a clickable drop-down list of all available concepts from the chosen framework. Hovering over the information icon provided details about each concept. If parents wanted to change the concept, they could select a new one from the list, triggering the LLM to automatically regenerate content based on the new concept (Figure \ref{fig:editor}).

\subsubsection{Activity Interface}
When an activity was launched, regardless of whether the parent had edited it, the \textit{book content} was presented on the right panel, with the \textit{LLM-generated content} and the \textit{modes-switching and role-delegation mechanisms} shown on the left. For each page, the components of the \textit{LLM-generated content} were organized into colored blocks, showing the \textit{book text}, \textit{questions and choices}, \textit{explanation}, and \textit{motivational feedback} in sequence. The \textit{mode-switching} mechanism enabled parents to control their involvement on a macro level by adjusting the overall task distribution for all components based on preset configurations. This mechanism used a driving metaphor, with positions for a driver, co-pilot, and exit. The parent was represented by a blue icon, and the robot by a green icon. By dragging the icons to the appropriate positions, parents could select their preferred mode for each task. The \textit{role-delegation} mechanism provided more detailed control, allowing parents to assign individual components of the activity to either themselves or the robot (Figure \ref{fig:activity}).

\begin{figure*}[t!]
  \includegraphics[width=\textwidth]{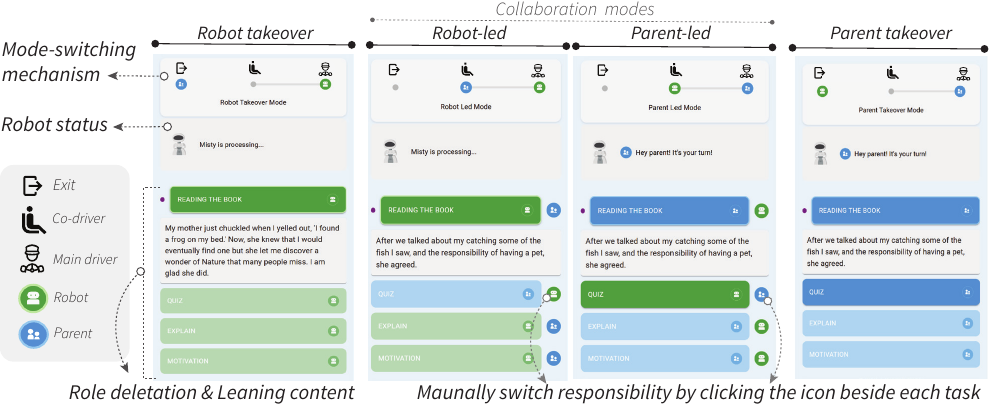}
   \vspace{-6pt}
  \caption{Mode-Switching and Role Delegation Mechanisms of the PAiREd system. The system offers four modes: robot takeover, robot-led, parent-led, and parent takeover. Parents can drag their icon to the ``driver,'' ``co-driver,'' or ``exit'' position to select the desired mode. Additionally, they can fine-tune the role delegation by assigning specific tasks to either themselves or the robot.}
  \label{fig:mode}
   \vspace{-6pt}
\end{figure*}

The system defined four modes: (1) \textit{parent-takeover} mode: the parent facilitates all content components; (2) \textit{parent-led} mode: the parent leads the activity, with the robot helping with specific tasks; (3) \textit{robot-led} mode: the robot leads, with the parent helping as needed; (4) \textit{robot-takeover} mode: the robot facilitates all components. The \textit{mode-switching} mechanism automatically adjusted the task roles based on the selected mode, assigning all roles to the parent in \textit{parent-takeover} and \textit{parent-led} modes, and all to the robot in \textit{robot-takeover} and \textit{robot-led} modes. In \textit{parent-led} and \textit{robot-led} modes, parents could further adjust the role distribution by delegating individual components using the \textit{role delegation} mechanism. Figure~\ref{fig:mode} details how icon positions correspond to specific modes and provides examples of a role delegation pattern for each mode.

   \vspace*{-3pt}
\subsection{Robot Interaction Design}
\begin{figure*}[b!]
  \includegraphics[width=\textwidth]{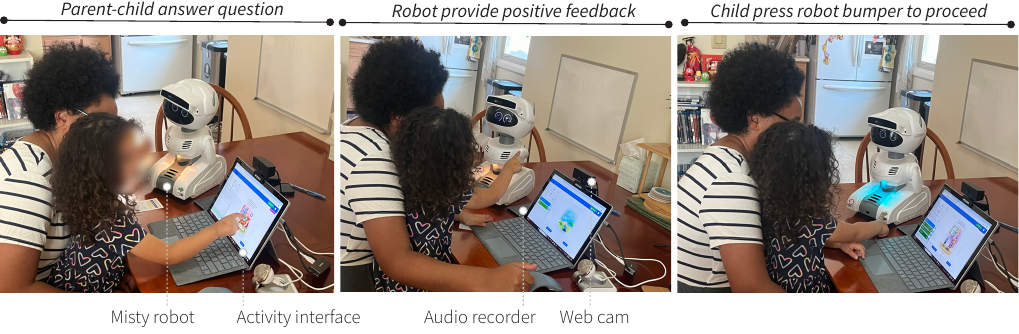}
   \vspace{-6pt}
  \caption{Robot Interaction Module of the PAiREd System and Study Setup. The robot facilitates the activity based on its assigned role, providing behavioral expressions to engage the child. The child can interact with the robot by pressing its bumper to navigate through the activity.}
  \label{fig:robot}
   \vspace{-6pt}
\end{figure*}

We used a Misty II social robot,\footnote{Misty Robot: \url{https://www.mistyrobotics.com/products/misty-ii/}} which is a semi-humanoid robot with a four-inch LCD display for its face where customizable facial expressions can be displayed. Misty robot was connected to our interface system through the Internet and the RestAPI.\footnote{\url{https://github.com/MistyCommunity/REST-API}} During the activities, the robot autonomously facilitated a component according to the role delegation, remaining silent when the parent was responsible for a component. The interface also displayed the robot's status and reminded parents when it was their turn to engage. For each component assigned to the robot, it verbalized the content using audio generated by Google's Cloud Text-to-Speech engine.\footnote{Google Cloud Text-to-Speech: \url{https://cloud.google.com/text-to-speech/}} Additionally, the system leveraged LLM to select an appropriate robot expressions per content component from a predefined database to enhance interaction through expressive gestures, which had been used in previous studies with children \cite{white2021designing}. The robot's front bumpers provided interactive controls, allowing young children to navigate the activity more easily. Pressing the left-front bumper \textit{repeated} the ongoing component, while pressing the right-front bumper \textit{proceeded} to the next component. When progressing to the next component, the interface updated accordingly by closing the current component and expanding the next one (Figure \ref{fig:robot}).

\begin{table}[t!]
    \caption{Participant Demographics}
    \label{tab:demographics}
    \centering
\fontsize{6.3}{8.3}\selectfont
    \begin{tabular}{p{0.02\linewidth}p{0.06\linewidth}p{0.09\linewidth}p{0.09\linewidth}p{0.09\linewidth}p{0.13\linewidth}p{0.23\linewidth}}%
        \toprule
         \textbf{ID} & \textbf{Child Age} & \textbf{Child Gender} & \textbf{Parent Gender} & \textbf{Ethnicity} & \textbf{Maternal Education} & \textbf{Household Income} \\
        \midrule
         1 & 3Y8M & Male & Female & Biracial & Master's & \$200,000 and more \\
         \midrule
         2 & 3Y4M & Female & Male & Biracial & Master's & \$150,000-\$199,999\\
         \midrule
         3 & 3Y2M & Female & Female & White & Master's & \$100,000-\$149,999\\
         \midrule
         4 & 4Y6M & Male & Female & White & Ph.D. & \$100,000-\$149,999\\
         \midrule
         5 & 4Y1M & Female & Female & White & Master's & \$100,000-\$149,999\\
         \midrule
         6 & 3Y3M & Male & Female & Black & Master's & \$35,000 -\$49,999\\
         \midrule
         7 & 3Y5M & Female & Female & White & Ph.D. & \$100,000-\$149,999\\
         \midrule
         8 & 4Y2M & Male & Female & White & Bachelor's & \$100,000-\$149,999\\
         \midrule
         9 & 3Y2M & Female & Female & Biracial & Ph.D. & \$50,000-\$74,999\\
         \midrule
         10 & 5Y0M & Male & Male & White & Bachelor's & \$100,000-\$149,999\\
         \midrule
         11 & 5Y1M & Male & Female & White & Ph.D. & \$200,000 and more\\
         \midrule
         12 & 4Y10M & Female & Female & White & Master's & \$100,000-\$149,999\\
         \midrule
         13 & 3Y9M & Male & Male & White & Master's & \$100,000-\$149,999\\
         \midrule
         14 & 4Y0M & Female & Female & Asian & Master's & \$50,000-\$74,999\\
         \midrule
         15 & 4Y6M & Female & Female & Biracial & Ph.D. & \$50,000-\$74,999\\
         \midrule
         16 & 4Y10M & Female & Female & White & Master's & \$150,000-\$199,999\\
         \midrule
         17 & 4Y7M & Female & Female & White & Master's & \$100,000-\$149,999\\
         \midrule
         18 & 4Y2M & Female & Female & White & Ph.D. & \$200,000 and more\\
         \midrule
         19 & 4Y9M & Female & Female & Hispanic & Bachelor's & \$200,000 and more\\
         \midrule
         20 & 3Y2M & Female & Female & White & Ph.D. & \$200,000 and more\\
         
         \bottomrule
    \end{tabular}
\end{table}

\section{User Study} \label{sec-study}

\subsection{Participants} 
Following protocols fully approved by the responsible Institutional Review Board (IRB), we recruited families from U.S. Midwest cities via email distributed to university employee mailing lists. We selected participants based on the following criteria: (1) one parent and one child participated together; (2) the child was aged 3--5 years; and (3) both parent and child could communicate in English. This age range was chosen as it represents a critical stage for parental involvement in early education \cite{purpura2013informal}, prior to formal schooling. Families received \$50 USD upon completing the study.

Our analysis includes data from 20 parent-child dyads with children aged 3–5 years (13 female, 7 male; $M = 4.1$, $SD = 0.67$ years). Each session involved one parent and one child, although other family members were sometimes present but did not participate. Participant demographics are summarized in Table \ref{tab:demographics}.
Most participating parents were mothers (17 mothers, 3 fathers), reflecting the greater likelihood of mothers serving as primary caregivers, consistent with previous research \aptLtoX[graphic=no,type=html]{\cite[{e.g.,}][]{schoppe2013comparisons}}{\cite[\textit{e.g.,}][]{schoppe2013comparisons}}. Despite efforts to recruit a diverse sample, the sociodemographic representation is limited: 85\% of mothers held at least a Master's degree, and 80\% of families reported an annual income of \$100,000 or more. This limits the generalizability of our findings and may omit insights into underrepresented groups, a limitation further discussed in Section~\ref{sec-7.4}.

\begin{figure*}[b!]
  \includegraphics[width=\textwidth]{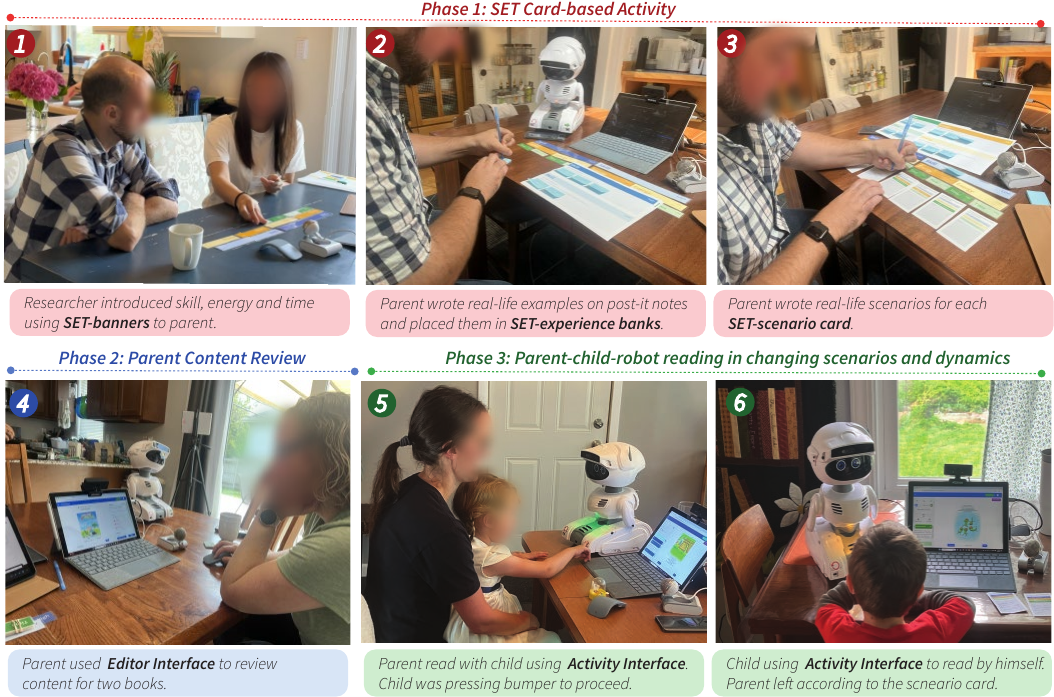}
   \vspace{-6pt}
  \caption{Three-phase study procedure. Phase 1 (40 min) and Phase 2 (40 min) only involved the parent, and Phase 3 (60 min) involved both the parent and the child. }
  \label{fig:procedure}
   \vspace{-6pt}
\end{figure*}

\subsection{Study Design} 
We conducted in-home visits (2.5 hours per visit) with families. We used \texttt{SET} (Section~\ref{sec-card} and Figure~\ref{fig:card-kit}) to foster discussion with parents about their real-life scenarios in which they are able or unable to facilitate learning activities for their child. In addition, we used \texttt{PAiREd}, an AI-assisted robot as a \textit{technology probe} \cite{hutchinson2003technology} to facilitate discussions on how parents may, under different \texttt{SET} scenarios, prefer to supervise the AI-generated content for their child and adjust their participation in learning activities.

\subsubsection{Study Materials and Setup}
The study materials included (1) a Misty II social robot; (2) a Microsoft Surface laptop to present user interface; (3) the \texttt{SET} card-based activity kit; (4) recording devices (\textit{i.e.,} a video camera, a webcam, and an audio recorder) positioned to capture participants' behavior and conversations (Figure~\ref{fig:robot}). 

\subsubsection{Study Procedure} \label{sec-procedure} 
Before the study, the experimenter provided families with an overview of the procedure and obtained informed consent: the parent signed a consent form, and the child gave verbal assent. The study was conducted in three phases: Phase 1 and Phase 2 (40 minutes each) involved \textit{only the parent} and Phase 3 (60 minutes) included \textit{both the parent and the child}. To minimize interruptions during the first two phases, the research team offered childcare assistance upon request (Figure \ref{fig:procedure}).

In Phase 1, the experimenter began by asking the parent to describe their involvement in their child's learning activities at home (\textit{e.g.,} ``\textit{What learning activities does your child typically do at home? Which ones involve you?}''). Next, the experimenter introduced the \texttt{SET} activity kit (Figure~\ref{fig:card-kit}, Section~\ref{sec-card}) and explained the factors and their dimensions with concrete examples (Table~\ref{tab:kit-example}), arranging the \texttt{SET-banners} into a hierarchical diagram for visualization (Figure~\ref{fig:card-kit}). The parent was then asked to write 2–3 real-life examples for each dimension of every factor on sticky notes and place them on the \texttt{SET-experience banks} (Figure~\ref{fig:card-kit}). Lastly, the experimenter introduced the \texttt{SET scenario cards}, explaining how the hierarchical diagram leads to eight scenarios by combining three two-level factors (Figure~\ref{fig:card-kit}). The parent was asked to write one real-life example for each scenario card for later use in Phase 3.

In Phase 2, the parent was first asked about their thoughts on AI-generated content (\textit{e.g.,} ``What are your thoughts on AI-generated content?''). The experimenter then introduced the \textit{editor interface} (Figure~\ref{fig:editor}), explaining its features for content modification. The parent reviewed and edited LLM-generated content for two books, one on math and one on literacy, while using the \textit{think-aloud} method to vocalize their thought process. Afterward, the experimenter discussed the parent's experience and used the \texttt{SET-scenario cards} to explore how they might use the interface in different scenarios.

In Phase 3, the parent first familiarized themselves with the robot interaction, learning to use the \textit{mode-switching} and \textit{role-delegation} mechanisms. The experimenter interviewed them about their perceptions of these mechanisms and their application in situations specified in \texttt{SET-scenario cards}. Next, the parent and child read two books together, incorporating activities created by the parent in Phase 2. At the start of each book, parents were asked to situate themselves in one of the \texttt{SET-scenario cards}, adjusting the mode and delegating jobs according to what they think they might do in that scenario. The experimenter gave the parent a different scenario card in the middle of the reading, and the parent responded to the situation, \textit{i.e.,} change to a different mode, accordingly. In addition, they also completed surveys on their perceptions of their child's math and literacy proficiency before and after the reading. Finally, they discussed their experiences collaborating with the robot and provided feedback on the designed mechanisms.

\begin{table*}
    \caption{Example of scenarios created using the {\texttt{SET-scenario cards}} by P2. P2 selected {\textit{``Teach child to build block towers''}} as the high-skill activity and {\textit{``Process feelings''}} as the low-skill activity for all examples. Each row represents one scenario card.}
\label{tab:scneario-example}
        \centering
        \renewcommand{\arraystretch}{1.2}
    \small
    \begin{tabular}{p{0.12\linewidth}p{0.12\linewidth}p{0.12\linewidth}p{0.12\linewidth}p{0.4\linewidth}}
        \toprule
        \hline
        \textbf{Scenario Card} & \textbf{Skill} & \textbf{Energy} & \textbf{Time} & \textbf{Example}\\
        \midrule
         1 & High & High & High & \textit{I am well rested and my partner is taking care of dinner.} \\
\hline
         2 & High & High & Low & \textit{Kids are being well-behaved but I have to mow the lawn.} \\
\hline
         3 & High & Low & High & \textit{I need to rest while my partner is making dinner.} \\
\hline
         4 & High & Low & Low & \textit{Kids are whining and I have to be at work.} \\
\hline
         5 & Low & High & High & \textit{Free from work and we don't have any plans.} \\
\hline
         6 & Low & High & Low & \textit{Well rested but have to do chores.} \\
\hline
         7 & Low & Low & High & \textit{Not feeling well but we have free time.} \\
\hline
         8 & Low & Low & Low & \textit{Not feeling well and have chores to do.} \\
\hline
         \bottomrule
    \end{tabular}
\end{table*}

\subsection{Data Analysis}
To anonymize participants, we assigned IDs based on the order of study completion (\textit{e.g.,} P1 refers to the first participant). For qualitative data, we conducted a reflexive \textit{Thematic Analysis (TA)} following \citet{clarke2014thematic} and \citet{mcdonald2019reliability} to identify patterns and understand parental perspectives. The first three authors, experienced in qualitative coding, transcribed and familiarized themselves with the data from audio and video recordings. Initial semantic codes were generated by the first author and organized into categories as the initial codebook. The team collaboratively coded data from P1 using the initial codebook, discussed interpretations, and created a consensus-based codebook. Using the codebook, the remaining data were independently coded by the three authors, with peer reviews and iterative discussions to finalize codes. We constructed final themes from recurring, meaningful patterns in the data. For the quantitative survey data (\textit{i.e.,} parent perception on child's literacy and math abilities), we first conducted a One-Way ANOVA to evaluate overall significance across difficulty levels for each subject (\textit{i.e.,} math and literacy), revealing significant differences between levels. We then performed Repeated Measures ANOVAs for each level, followed by post-hoc pairwise comparisons using non-parametric (Wilcoxon Signed-Rank) tests.
\section{Results}
We identified key contexts of parental involvement, perceptions of AI-generated content, preferences for AI-assisted content creation, and collaborative patterns in shared interactions with a robot. We present the findings based on our research questions as follows.


\subsection{(RQ1) What contexts do parents encounter when involving in young children's learning activities?}

Each participant provided examples for the eight scenarios representing unique combinations of the three two-dimensional factors using the \texttt{SET-scenario cards}. We present P2's scenarios as an example in Table~\ref{tab:scneario-example}. The full set of scenario examples from all participants is documented as supplementary materials.\footnote{SET Scenarios: \url{https://osf.io/zfksg/?view_only=b59bd41287f543ce82ab85950aaf004f}} Beyond capturing the contexts shared by each parent, we analyzed these examples to identify and summarize key contextual patterns for the three factors of parental involvement: \textit{skills}, \textit{energy}, and \textit{time}.

\subsubsection{\textbf{Skill:} Parents face challenges in pedagogical skills, particularly with advanced or unfamiliar concepts.}
Parents mentioned several \textit{skills} in supporting their child's (1) intellectual, (2) pedagogical, and (3) social-emotional development, highlighting key challenges across these areas. While some parents (7/20) reported low confidence in \textit{intellectual} activities, especially advanced STEM topics (P1, P7, P12, P14, P17–19), many (16/20) felt confident in literacy (\textit{e.g.,} reading, spelling; P1–6, P8, P9, P11, P13, P14, P19, P20) and basic STEM (P6, P8–10, P12–15). Confidence often stemmed from personal expertise or interests, consistent with the Hoover-Dempsey and Sandler (HDS) framework \cite{green2007parents}. For example, P15, a physicist, felt confident teaching physics-related activities. The majority of parents (14/20) struggled with \textit{pedagogical} skills, such as explaining concepts (P7, P8, P13, P17, P18), answering or formulating questions (P3, P4, P6, P7, P9, P13), identifying developmental benchmarks (P4, P6, P10, P11), and allowing their child to learn from mistakes (P2, P12, P19). A smaller group of parents (7/20) expressed confidence in these areas, particularly explaining concepts (P14, P18, P19) and answering questions (P4, P5, P10, P14). \textit{Social-emotional} skills presented additional challenges. Some parents (6/20) struggled with teaching emotion regulation (P2, P17), behavioral management (P5, P15, P20), and interpersonal conflict resolution (P3, P15). Others (5/20) lacked confidence in encouraging participation in learning activities (P5, P11) or maintaining patience during learning support (P4, P10, P18, P19). Conversely, several parents (10/20) felt confident teaching emotion regulation (P1, P3, P10–13, P16, P18–20) and norms of polite communication (P7, P12, P16, P20).

\subsubsection{\textbf{Energy:} Parents' motivation depends on their physical and emotional status as well as the child's willingness to learn.}
Parents suggested that their motivation to facilitate learning activities was affected by (1) physical status, (2) emotional status, and (3) time. Commenting on their \textit{physical status}, most parents (16/20) indicated low motivation when they need rest due to feeling ``\textit{hungry},'' ``\textit{sick},'' or ``\textit{tired}'' (P1--11, P14--17, P19), and many parents (9/20) reported being highly motivated when they are ``\textit{well rested}'' or after having ``\textit{a really good meal}'' (P1--4, P6, P7, P9, P11, P16). Regarding \textit{emotional state}, many parents (9/20) lacked motivation when they needed a mental break or ``\textit{me time}'' if they felt emotionally exhausted (P3, P7, P11--14, P17--19) or after spending time with their child (P5, P8, P20). In addition, some parents (7/20) lost motivation if their child appeared to be disinterested (P11, P12, P14) or poorly behaved (P4, P15, P17, P20). In contrast, many parents (16/20) were motivated when their child needed support (P12, P15), expressed interest and invited the parent to participate (P1, P4, P7, P8, P10, P12–14, P16–19), or is well behaved and ready to learn (P2, P5, P13, P18, P19, P20). Some parents (6/20) were highly motivated when they wanted to connect with their child (P3, P14, P15) or when they were personally interested in the activity (P5, P10, P12, P14).

\subsubsection{\textbf{Time:} Parents' availability depended on work, chores, other family members.}
Parents discussed (1) work and commitment, (2) household chores, and (3) family needs as factors that determined whether they had time, \textit{i.e.,} availability and presence, to facilitate learning activities. Most parents (19/20) were not available when they needed to be at \textit{work} (P1--4, P6, P7, P10--14, P16--19) and had other personal or professional engagements (P4, P5, P9, P11, P15, p20). Many parents (17/20) stated that \textit{household chores}, such as laundry, meal preparation, and cleaning, also determined their availability to be with their child (P1--6, P9--17, P20). Although some parents involved their child in chores (P1, P2, P7, P10, P15, P17), not all chores were seen as being appropriate or safe for children. Parents' availability also depended on the ability of other family members to provide support (P2--4, P6--8, P16--20), \textit{e.g.,} when a spouse helped with chores or an older child watches a younger sibling. Parents had less time if other family members needed them (P3, P7, P8, P12, P13, P16--18, P20), \textit{e.g.,} when a younger child is crying or a family member is sick. Finally, parents described their availability using specific time frames, \textit{e.g.,} ``\textit{weekday mornings} (P5, P8, P19),'' ``\textit{weekdays after dinner and before bedtime} (P5, P10, P18),'' ``\textit{anytime on weekends} (P1, P3, P4, P6, P8, P10, P16),'' or ``\textit{unstructured time} (P2, P11, P14, P16, P17, P20).'' They often structured their time and consider themselves available when they are physically present with their child (P1, P3--7, P14--16), such as during grocery shopping, car rides, or trips to the park together.

\subsection{(RQ2) How do parents perceive AI-generated content for young children?}\label{sec-result-2}

Parents showed mixed attitudes toward AI-generated learning content for young children. They discussed their perceived benefits and risks and envisioned ways to mitigate their concerns.

\subsubsection{Mixed Attitude towards AI-generated content}
Parents expressed a range of attitudes towards allowing AI to generate content for young children, ranging from skepticism and concern (P2--6, P10, P13--15) to open-minded caution (P9, P8, P16, P19, P20), acceptance (P6, P7, P11, P17, P18) and, in some cases, neutral (P1, P12). Parents who were \textit{\textbf{skeptical and concerned}} questioned whether AI-generated content met quality and safety standards, \textit{e.g.,} P3 questioned, ``\textit{Who's generating the content? Where is it getting the content from? Is it good? Is it safe?}'' On the other hand, parents who hold an attitude of \textit{\textbf{open-minded caution}} recognize the risks of using AI-generated content but feel open to use it under specific conditions. P16 highlighted model training, stating, ``\textit{I wouldn't be against it if the people training it were proficient in what the AI is teaching.}'' Similarly, P20 emphasized personal oversight, explaining, ``\textit{I can do my own evaluation to determine whether or not I think the content is good regardless of who it came from.}'' Furthermore, parents who have an attitude of \textit{\textbf{acceptance}} assume people who created the system have already ensure the appropriateness for children, \textit{e.g.,} P7 stated, ``\textit{I'm assuming because it's AI, there would be more research behind it.  So I would be okay with it.}'' Finally, parents who hold a \textit{\textbf{neutral}} attitude typically don't have much experience with AI and therefore feel unsure about their attitude for AI-generated content, \textit{e.g.,} P1 had ``\textit{not even thought about it until before this study.}''

\subsubsection{Perceived Benefits and Risks}
Parents identified several benefits of AI-generated content for young children. Some parents (P2, P4, P8, P10, P11, P16) highlighted AI's potential in \textit{\textbf{adaptability}} to adjust learning content to their child's evolving developmental needs, \textit{e.g.,} P11 expected AI to help ``\textit{adjust content as the child grows.}'' In addition, parents (P2, P3, P4, P16, P19) discussed \textit{\textbf{customization}}, illustrating that ``\textit{one of the big benefits would be to create material that are related to his[child's] interests and things that would be motivating to him[child]} (P4).'' Parents (P6, P7, P8, P10, P12, P17) also emphasized \textit{\textbf{efficiency}} of AI, explaining ``\textit{because it[AI] can access a huge amount of information very fast} (P12),'' enabling a ``\textit{quicker way to learn or to see something} (P7).'' Moreoever, a few parents (P1, P11, P18, P20) noted AI's potential to foster \textit{\textbf{affordability}}, suggesting that AI-generated content could enhance the scalability and accessibility of learning resources, making ``\textit{more learning materials available, more variety available} (P1),'' and making things ``\textit{cheaper and more accessible for people} (P20).'' Finally, a few parents (P14, P15) expected easier \textit{\textbf{pedagogical integration}} with AI, enabling parents to ``\textit{teach children things that sometimes parents don't know because not all parents know everything} (P14).''

Meanwhile, parents described their perceived risks of AI-generated content for children. Most parents (P1--3, P5, P11--15, P17, P19) were concerned about \textit{\textbf{age-inappropriateness}} of the content, which could be ``\textit{violent and don’t match family values} (P1),'' ``\textit{physically harmful and sexually inappropriate} (P3),'' and ``\textit{stuff about body image and certain people being better than other people} (P5).'' In addition many parents (P2, P3, P9, P14, P16, P17, P18) expressed concerns about the \textit{\textbf{inaccuracy}} of the information presented through AI-generated content, worrying that AI could provide ``\textit{factually inaccurate}'' learning materials or content that might imply theories that are ``\textit{misframed or misconstructed} (P2).'' Moreover, parents (P2--4, P14, P15) raised concerns about the \textit{\textbf{training data quality}} for AI models. P2 emphasized transparency stating, ``\textit{I'd want to know a lot more about where that training data came from or who supervised that learning process}.'' A few parents (P6, P7) expressed concerns about children's \textit{\textbf{over dependence}} on AI instead of developing their own cognitive abilities. P6 worried that constant use of AI could discourage critical thinking, stating, ``\textit{if they have a question, instead of thinking through the question, they just ask AI, not using their own brain}.'' Finally, two parents shared concerns over \textit{\textbf{message dilution}}, where AI oversimplifies complex ideas and diminishes their original intent. P15 worried that AI might dilute sociopolitical issues, such as racial diversity and gender identity. Similarly, P20 emphasized concern about whether the core message being conveyed to the child aligns with parental values, stating ``\textit{I'm more concerned about the message the book is trying to impart on the child}.

\subsubsection{Envisioned Risk Mitigation Methods}
Parents described what methods they envisioned to address their concerns. First, some parents (P3, P5, P11, P12, P14, P16) stressed the need to enable \textit{\textbf{parental review and verification}}. For example, P5 stated ``\textit{I would read it to make sure that it was actually something I wanted to read with her}.'' In addition, a few parents (P2, P17, P19) expressed that \textit{\textbf{social and public validation}} could also enhance their trust in AI-generated content, \textit{e.g.,} P2 described that ``\textit{if a thousand people used it...and endorsed this model, that would give me more confidence in it}.'' Moreover, some parents (P2, P9, P15) discussed \textit{\textbf{model and data transparency}}, emphasizing the need to understand how AI models are trained. As explained by P9, ``\textit{being able to know exactly what's going on or how it works...would make me feel more secure about what my child is learning}.'' Lastly, a few parents (P1, P15, P18) highlighted the importance of \textbf{expert involvement} in creating AI-generated content. For instance, P1 emphasized the need for oversight by ``\textit{people with a background in human development.}''

\subsection{(RQ3) How would parents prefer to collaborate with LLM on supervising content creation under different contexts?}

\begin{figure*}[b]
\includegraphics[width=\textwidth]{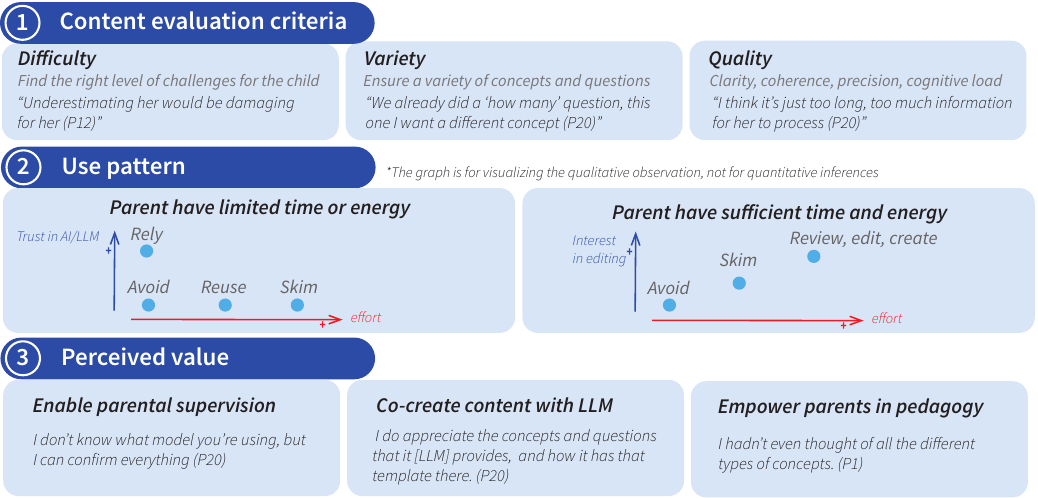}
   \vspace{-6pt}
  \caption{Summary of parent's use of LLM-assisted content supervision mechanism: (1) content evaluation criteria, (2) use pattern, (3) perceived value.}
  \label{fig:result-03}
   \vspace{-6pt}
\end{figure*}

We found three main themes for parent-AI collaboration on content creation using the \textit{editor interface}: (1) \textit{Content evaluation and criteria}, referring to what parents pay attention to when reviewing and revising LLM-generated content. (2) \textit{Contextual usage patterns}, describing how parents envision using the LLM-powered interface in various contexts. (3) \textit{Perceived value and benefits}, covering what values parents believe LLM brings.

\subsubsection{Theme 1: Content evaluation and criteria}
Parents focus on balancing \textit{difficulty} and \textit{variety} of concepts as well as ensuring the \textit{quality} of questions when reviewing, regenerating, and revising LLM-generated content for young children.

\textit{\textbf{Parents aim to give their children the right level of challenge while reinforcing skills they can confidently accomplish.}} Many (8/20) avoided overly easy questions to prevent boredom but strategically included them at the start or after difficult questions to build confidence. As P2 explained, ``\textit{I want her to get the answers and then have it get increasingly difficult as she goes so she doesn't get discouraged at the beginning}.'' Meanwhile, most parents (11/20) valued challenges that stretch their child's abilities without overwhelming them. P12 concerned that ``\textit{underestimating her would be damaging for her},'' while P20 expressed interest in seeing how his child would handle harder concepts, saying, ``\textit{I'm actually really interested to see if she can answer.}'' Finally, parents (7/20) were also cautious of content that might be too advanced, \textit{e.g.,}``\textit{she[child] doesn't know uppercase or lowercase yet, so that doesn't mean anything to her} (P20).'' Additionally, \textit{\textbf{parents aim to maintain engagement by introducing diverse concepts and question types throughout the activity.}} Many parents (11/20) expressed concerns over repetitive content and preferred diverse topics to challenge their child differently. For instance, P20 changed the concept of a question to ``addition'', explaining, ``\textit{I just made the last question a `how many,' so this one I want a different concept}.'' Finally, \textit{\textbf{parents evaluate the quality of LLM-generated learning content based on standards} such as question clarity and coherence (9/20), wording precision (6/20), visual clarity (5/20), and cognitive load (P12, P20).} P12 raised issues with wording, stating, ``\textit{I don't think she's going to fully understand front legs versus back legs when it's a front view},'' while P6 expressed concerns about visual clarity: ``\textit{from the pictures, you can't really tell how many bugs with black bodies are flying in the air}.'' P20 also reflected on cognitive load, saying, ``\textit{I think it's just too long, too much information for her to process}.'' 

\subsubsection{Theme 2: Contextual usage patterns}
We discussed parents' preferences and behaviors when collaborating with LLM under two main contexts: (1) when parents have limited time or energy and (2) when they have sufficient time and energy.

\textit{When parents have limited time or energy}, most were still \textit{willing} to invest minimal effort (P4, P6--9, P11, P12, P15, P18), often opting to \textit{\textbf{skim through the LLM-generated content with minor self-editing}}. For example, P9 shared, ``\textit{I might skip quickly, skim through it, make sure there isn't anything that I feel is not appropriate}.'' This approach allows involvement with minimal time commitment. However, some parents (P4, P11, P15) emphasized that the \textit{\textbf{LLM output must be high-quality enough to require minimal editing}}, otherwise they may not use it at all. P11 explained, ``\textit{The more that stuff can be in really good shape before it gets to parents, the more we can minimize how much work we have to do ahead of time}.''

Some parents were \textit{unwilling} to invest effort when time or energy was limited. They preferred to either \textit{\textbf{reuse previously reviewed activities}} (P8, P11, P12) or directly \textit{\textbf{use LLM-generated content without review}} (P5, P6, P9, P10, P18, P20). As P8 explained, ``\textit{if I don't have time, I would have to be using something he's already done before, so I don't have to supervise it},'' while P10 noted, ``\textit{if the AI-generated questions were enough to keep him engaged, then it would be worth it}.'' A few parents (P1, P7) preferred to \textit{\textbf{avoid using the system entirely}}, as they feel uncomfortable leaving their child engaged with the content without supervision. As P1 explained, ``\textit{if I'm either physically or mentally not present. It's just not happening}.''

\textit{When parents have sufficient time and energy}, most of them (P6–P10, P12, P18, P20) choose to \textit{\textbf{review and edit the content in detail, even customizing questions}} to better supervise and personalize learning for their child. P9 shared, ``\textit{If I had more time and motivation, I would take the time to do it myself. I enjoy writing, so I'd probably spend time customizing the content}.'' Similarly, P12 noted, ``\textit{If I had all the time, I would go through and be picky with the wording and content of the questions}.''

In contrast, some parents (P1, P2, P7, P11, P12, P18) still prefer to \textit{\textbf{skim through the content with minor editing}}, as they found the detailed process too effortful even when time allowed, but they cannot fully trust LLM or themselves to come up with good questions. For example, P11 shared, ``\textit{I would probably scroll through and try to do as little editing as possible},'' while P7 expressed doubt, stating, ``\textit{I don't know that I would come up with better questions than this one from AI}.'' A few parents opted to \textit{\textbf{avoid using the system entirely}}, preferring to spend their time on other activities (P15) or relying on their ability to engage their child without the system (P4, P5). For instance, P15 shared, ``\textit{I would rather spend that time playing an imaginative game with her than spending time designing this,}'' P5 similarly expressed confidence, saying, ``\textit{I think I can and do ask her questions about stuff we read}.''

\subsubsection{Theme 3: Perceived value and benefits}
We found that parents perceive the value of the system to include not only \textit{content supervision}, but also \textit{content co-creation with LLM} and \textit{parent empowerment through pedagogical insights}.

First, and unsurprisingly, most parents suggested that the system allows them to \textit{\textbf{supervise the learning content generated by LLM}}. For example, P2, while feeling skeptical about trusting AI, noted, ``\textit{I don't know what AI model was used, still, I can confirm everything myself},'' reflecting the value parents place on maintaining oversight of the content presented to their children. Second, some parents (P6–10, P12, P18, P20) appreciated that the system allows them to \textit{\textbf{co-create personalized learning content with the LLM}} for their child without having to start from scratch. For example, P18 appreciated the ability to adapt the content to their child's needs, saying, ``\textit{tailoring it to her difficulty levels and seeing the ability to modify the content alleviates some concerns}.'' P10 highlighted how the LLM creates a draft to work from, stating, ``\textit{I do appreciate the concepts and the kinds of questions that it [LLM] provides, and how it has that template there}.'' This flexibility allowed parents to easily modify content while leveraging the assistance from LLM. Third, some parents found that the system \textit{\textbf{empowered parents with pedagogical insights}}. As many parents do not possess formal pedagogical knowledge--such as understanding how to effectively teach their child--they often struggle with determining what questions to ask or which concepts are age-appropriate. Since parents brought up the same value after interacting with the robot as well, we discuss this value more in-depth in Section \ref{sec-6.4.2}.

\subsection{(RQ4) How would parents prefer to collaborate with an AI-assisted robot to engage in learning activities with their children under different contexts?}

\begin{figure*}[b!]
\includegraphics[width=\textwidth]{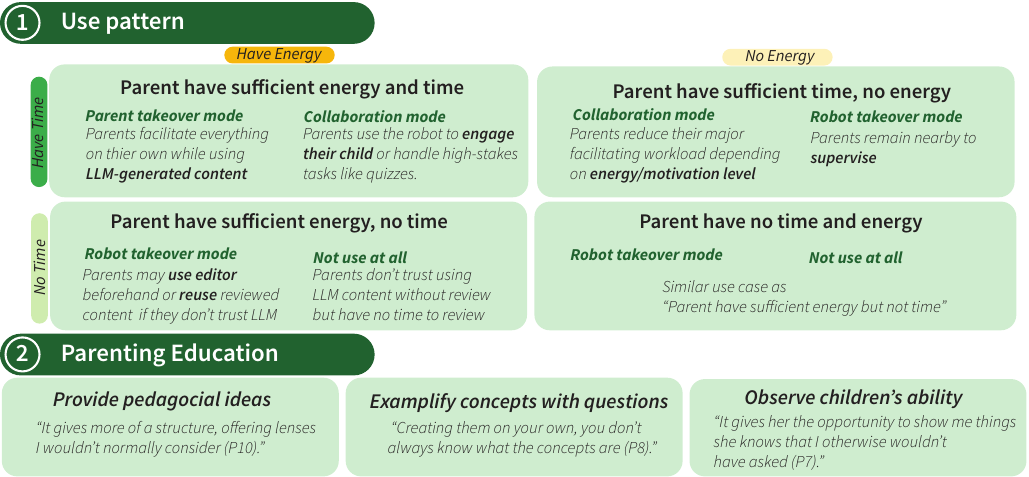}
   \vspace{-6pt}
  \caption{Summary of parent's use of robot involvement adjustment mechanisms: (1) usage pattern, (2) parenting education.}
  \label{fig:result-03}
   \vspace{-6pt}
\end{figure*}

We identified two major themes in the use of parent-robot collaboration mechanisms (\textit{i.e.,} \textit{mode-switching} and \textit{role-delegation}) within the \textit{activity interface}: (1) \textit{Contextual mode utilization}, referring to how parents adjust their involvement based on varying time and energy levels, and (2) \textit{Perceived educational impact on parenting}, highlighting how parents value the process for enhancing their skills and knowledge in parenting.

\subsubsection{Theme 1: Contextual mode utilization}

We discussed parents' preferences when collaborating with the AI-assisted robot across four contexts: (1) sufficient energy and time, (2) sufficient time but low energy, (3) sufficient energy but limited time, and (4) low energy and time. The impact of parental skill is discussed in specific cases.

(1) \textit{Parents have sufficient energy and time}: many parents (P2, P5–7, P9, P15, P18) preferred the \textit{\textbf{parent takeover mode}}, where they facilitate activities themselves while using LLM-generated content as a resource. For example, P18 shared, ``\textit{if I'm feeling motivated, I'd probably take over, but still look at some AI-generated questions to prompt me or remind me of things to ask or do with her}.'' Similarly, P15 noted, ``\textit{with full energy and time, I would use the parent-only mode because I want to interact with her and give her all my attention}.'' Parents valued the ability to take full control while using LLM-generated content for supplemental support when they have sufficient energy and time.

In addition, some parents (P2, P8–12, P15, P20) envisioned using \textit{\textbf{collaboration modes}}--where both the parent and the robot share responsibilities (\textit{i.e.,} parent-led or robot-led mode)--with the parents' \textit{skills} in specific areas relative to the robot playing a critical role in determining the pattern of role delegation. Parents often chose to involve the robot when they felt it could enhance their child's engagement especially in high-stakes tasks like quizzes. For example, P15 noted, ``\textit{I would use the robot for quizzes as a playful element to keep her engaged}.'' Similarly, P2 highlighted the objectivity of the robot in quizzing: ``\textit{I like the idea of reading her the book and then a neutral third party gets to test her on it}.'' On the other hand, parents took on specific roles when they believed their involvement would benefit their child more. P11 shared, ``\textit{I would let the robot read and ask questions but step in if he wasn't understanding or needed guidance},'' while P12 emphasized the emotional aspect of teaching: ``\textit{I can explain in a way that she understands, whereas the robot might come across as too harsh}.''

(2) \textit{Parents have sufficient time but lack energy}: some parents (P3, P7, P10--12, P15, P18, P20) opted for \textit{\textbf{collaboration modes}}, with their involvement influenced by their motivation levels and partially by their \textit{skill} relative to the robot. For example, P10 noted, ``\textit{when I'm not motivated, having the robot do the quiz takes some heat off me}.'' Similarly, P9 mentioned, ``\textit{I'd probably read the book, but have the robot do everything else}.'' Additionally, some parents (P1, P2, P4, P6–8, P15) chose to use \textit{\textbf{robot takeover mode}}--where the robot facilitates everything--while they remained nearby to supervise. For instance, P15 noted, ``\textit{I'd be around, but I wouldn't physically do much because I'm not feeling well}.'' Similarly, P2 noted, ``\textit{If I'm not motivated, I could see myself handing it all over to the robot}.''

(3) \textit{Parents have sufficient energy but lack time}: many parents (P2, P3, P5, P7--11, P15) opted for the \textit{\textbf{robot takeover mode}}--where the robot facilitates everything--while adjusting their usage based on \textit{how much they trust LLM}. Parents with higher trust allowed their child to use the content directly without review (P3, P5, P7, P9, P10). For example, P7 mentioned, ``\textit{If I'm trying to take a walk, I might do the robot takeover, then I can physically be gone}.'' In contrast, parents with less trust preferred to supervise while multitasking (P2, P7, P8), review content beforehand using the editor (P2, P8), or use the system only if the LLM model met high-quality standards (P11, P15). For example, P2 shared, ``\textit{If I'm not there, I wouldn't want them to do it, unless I had used the editor to review},'' while P7 described a multitasking scenario: ``\textit{I could be working from home while the robot takes over, and I'm nearby to supervise}.'' Moreover, a few parents (P2, P12) chose to \textit{\textbf{avoid using the system entirely}} due to their lack of trust in using LLM-generated content directly and insufficient time to review it, or because the system design did not support independent use for young children (P4, P12, P20). For instance, P2 noted, ``\textit{If I'm absent, I don't know if I'd want them to do any of this},'' while P12 stated, ``\textit{I know my daughter is sensitive, and if [the questions are too hard and] the robot keeps telling her she's wrong, she might take it personally and give up}.''

(4) \textit{Parents lack both energy and time}: Some parents chose \textit{\textbf{robot takeover mode}}, adjusting their usage based on their trust in LLM-generated content. Others \textit{\textbf{avoided using the system entirely}} due to low trust and insufficient time to review (P2, P12), or because the system design did not support independent use by young children (P4, P12, P20). Refer to the previous case—parents with sufficient energy but lacking time—as the usage patterns and contextual reasons are very similar.

\begin{figure*}[!t]
  \includegraphics[width=\textwidth]{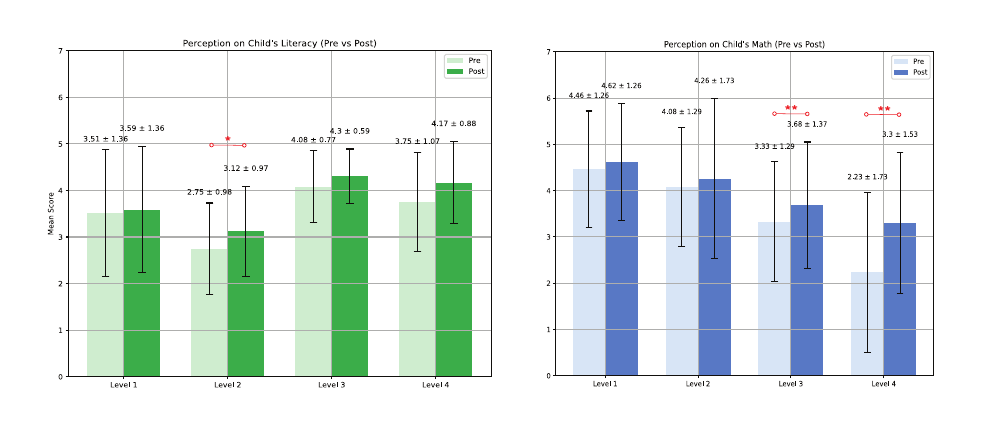}
  \caption{Parent perception on child's math and literacy ability before and after the reading session. The result suggested that parents adjusted their perception after observing their child doing the activity and they tend to underestimate them, especially for advanced math concepts and phonological awareness concept for literacy. The horizontal lines represents significance from the Wilcoxon Signed-Ranked Test: $p < .01^{**}$, $p < .05^{*}$.}
  \label{fig:quant-result}
   \vspace*{-10pt}
\end{figure*}

\subsubsection{Theme 2: Perceived Educational Impact on Parenting} \label{sec-6.4.2}

Supported by mixed-method data, many parents thought \texttt{PAiREd} has value in parenting education, providing them with pedagogical strategies and giving them opportunities to observe their child's proficiency level systematically through observation. If the system provides a comprehensive framework and ample ideas, parents may not decrease their involvement in an activity just because they don't have the pedagogical skill; in fact, they may even increase their involvement. In addition, parents will be able to observe and adjust their understanding about what their child can do or cannot do, instead of under- or over- estimate their child's ability.

Several parents (P1, P4, P6, P7, P10--12, P18) appreciated that the system \textit{\textbf{offered ideas they might not have considered on their own}}, providing new topics to explore with their child. For instance, P1 emphasized, ``\textit{I hadn't even thought of all the different types of concepts},'' and P10 highlighted that the system ``\textit{gives more of a structure…even the drop down list of concepts is insightful, offering lenses I wouldn't normally consider when reading}.'' Others (P2, P6--9, P11, P12) valued that the LLM \textit{\textbf{generated example questions for each concept}}, allowing them to start with ready-made content without worrying whether their own questions reflected the intended learning goals. For example, P8 mentioned, ``\textit{What's nice about the AI-generated ones is that you can specifically choose a variety of concepts, whereas creating them on your own, you don't always know what the concepts are}.'' Additionally, some parents noted that the system allowed them to systematically select questions they were unsure their child could answer, which \textit{\textbf{provided a structured way to observe and assess their child's proficiency level}}. For example, P7 remarked, ``\textit{it gives her the opportunity to show me things she knows that I otherwise wouldn't have asked},'' while P8 shared, ``\textit{I was curious to see how he does if he doesn't know how to answer this, rather than just setting him up to succeed}.''

Before and after parent-child pairs engage in the activity, we asked parents to rate their perception on their child's math and literacy abilities. Our quantitative results suggest that parents adjusted their understanding of their child's proficiency in certain concepts after reading together. Specifically, \textit{\textbf{parents tended to underestimate what their child can or cannot do, especially with more advanced math concepts}} (Math-L3: $p < .01^{**}$, Math-L4: $p < .01^{**}$) and the phonological awareness concept in literacy ($p < .05^{*}$). Figure \ref{fig:quant-result} summarizes the significance results from the Wilcoxon Signed-Rank Test.


\section{Discussion}
Below, we draw upon our findings to explore the idea of enabling parents to collaborate with an AI-assisted robot to supervise generative learning materials and flexibly involve in their child's learning.

\subsection{Parental Involvement Support Needs}
We found that while parents generally have the \textit{skill} to support their children's development, they seek more guidance on pedagogical strategies, especially for advanced concepts. In addition, our quantitative results further indicate that parents tend to underestimate their children's ability in more advanced concepts. In fact, \citet{claessens2014academic} and \citet{engel2013teaching} suggested that preschoolers benefit from exposure to advanced content in literacy and math, compared to basic content. Parents' lack of ability or confidence to guide children in advanced content hinders more effective support for children. Therefore, supporting parents with pedagogical strategies in advanced content and providing them with opportunities to observe their children's developmental progress can further optimize the value of parental involvement. In addition, we discovered that parents' \textit{motivation} to engage in learning activities is influenced by their physical and emotional state and their children's willingness to learn, while parental \textit{availability} is shaped by factors such as work, household responsibilities, and family needs. These results exemplify prior parental involvement frameworks \cite{green2007parents, ho2024s}, and future studies should consider how to detect these moments reliably to automate intelligent decisions to support parents.

\subsection{Ethical Consideration in the Use of AI-Generated Learning Content}\label{sec-dis-2}
We found that parents have mixed attitudes toward AI-generated learning content, expressing concerns about age appropriateness, inaccuracies, data quality, over-reliance, and message dilution. Similar concerns were raised by \citet{han2024teachers}, who noted parents' challenges in distinguishing elementary-school students' authorship and controlling misinformation. \citet{yu2024exploring} and \citet{ho2024s} also highlighted worries about misinformation, appropriateness, and privacy for teenagers. To mitigate harm from AI for children, prior research has outlined five key principles for age-appropriate AI: fairness and equality, transparency and accountability, privacy and exploitation, safety and safeguarding, and sustainability and age-appropriateness \cite{wang2022informing}. These insights not only underscore the principles AI-generated content should adhere to but also emphasize parents' concerns about the use of such content for children. Researchers should continue developing principles to guide generative AI for children while also proposing effective interaction paradigms that facilitate parental supervision. Furthermore, enhancing the explainability and transparency of how learning materials are generated by AI could help alleviate parental concerns and encourage more effective use. While creating AI models that produce safe and accurate outputs is crucial, designing mechanisms that build parents' trust and understanding of the content's source could improve the adoption of AI-generated materials.

\subsection{Optimizing Parental Control and Supervision for LLM-Generated Content}
We identified three key factors--difficulty, variety, and quality--that parents considered when reviewing LLM-generated learning content, providing essential criteria for developing specialized AI models for early education. In addition, many parents expressed an interest in leveraging content generated by AI but also voiced concerns, consistent with prior studies on parental control over media content \cite{yu2024parent, ho2024s} and preferences for creating their own learning materials using AI \cite{han2024teachers}. We found that parents' willingness to control and supervise the content varied by their \textit{level of trust in LLMs} and \textit{personal preferences}. When time or energy was limited, many parents opted for minimal supervision, skimming or directly relying on LLM-generated content. Those with lower trust tended to rely on previously reviewed materials or avoid the system entirely. Even when parents had more time and energy, their approaches diverged--some enthusiastically customized the AI-generated content, while others viewed customization as a burden and minimized or avoided using the system.

The range of use cases indicate that, first, the questions auto-generated by LLMs need to meet parents' expectations for quality and safety. This would allow parents to trust and make minimal edits, thereby improving their confidence in LLMs. Second, to further reduce parental effort, especially when time and energy are limited, LLM-based content generation should adapt by learning from parents' and children's preferences over time, minimizing repetitive errors. Third, the design for varying levels of control over LLM autonomy should be refined to suit different use cases. For example, a flexible design could allow parents to toggle between reviewing only critical information and in-depth editing. Future research and design should prioritize improving the quality of LLM-generated content, develop mechanisms for LLMs to incorporate parental feedback over time, and design user-friendly interfaces that provide flexible control options, reducing cognitive load across different use scenarios.

\subsection{Design for Complex and Contextualized Interaction Dynamics with AI-based Robot} 
We examined parents' contextual usage patterns of the editor interface, revealing a spectrum of approaches: editing, skimming, reusing previously reviewed content, using LLM-generated content directly, or avoiding the system altogether. Similarly, in the activity interface, we observed various usage patterns: \textit{parent takeover} mode leveraging LLM content as support; \textit{collaborative} modes determined by parent skill and motivation; \textit{robot takeover} mode under supervision, with previewed content, or with original LLM content; and opting to avoid the system entirely. \citet{zhang2022storybuddy} similarly emphasized the importance of flexible parental involvement during reading through a system called \textit{Storybuddy} using a virtual chatbot, while \citet{dietz2024contextq} presented auto-generated dialogic questions to caregivers for dialogic reading that is similar to our \textit{parent-takeover} mode. 

The parent-child-robot interaction must adapt to changing scenarios. For example, when a parent becomes busy or loses motivation, the dynamic may shift from parent-child-robot interaction to child-robot interaction. To adapt to this shift, designing mechanisms such as mode-switching and flexible role delegation is essential. This approach allows parents and children to utilize the robot across a broader range of contexts, rather than being restricted to specific scenarios, thereby catering to individual needs and preferences. Although we made an initial attempt to design a mechanism that enables mode-switching and flexible role delegation between the parent and the robot, our findings revealed areas for improvement. We also identified factors that can hinder parents and children in utilizing this design to its full potential. 

First, when parents choose the \textit{parent takeover mode}, they often do not require the robot at all. Instead, they rely on LLM-generated content as a resource to guide their conversations, pacing the interaction according to their own and their child's needs. The current presentation of information may not optimally support parents in leading these activities.  Thus, the \textit{parent takeover mode} should focus on delivering suggested interaction content more efficiently, enabling parents to access and integrate the material seamlessly. P4, for example, found this mode challenging because her child was too captivated and distracted by the robot, reducing her ability to engage fully. This observation suggests that the design should allow parents to easily separate the robot from the content delivery, providing them with full control. From an educational standpoint, since parents are deeply involved in this mode, it presents opportunities to introduce more advanced content. Parents can explain complex concepts in a manner tailored to their child’s understanding compared to the robot. Therefore, integrating or suggesting more advanced content in this mode could enhance parents' engagement and maximize the educational value of their involvement.

Second, when parents select \textit{collaboration modes} (either parent-led or robot-led), they often relied on the robot to either engage their child or reduce their own involvement when feeling tired. 
These different needs--enhancing engagement versus offloading activity leadership--require tailored design solutions. For instance, the robot could not only facilitate learning but also offer entertaining and interactive responses. Parents could activate these features as needed, drawing attention to the content or creating excitement.

Third, when parents selected the \textit{robot takeover mode}, they wanted to either reduce their involvement while maintaining supervision or be completely absent. In the first case, they preferred the robot to lead the activity while they remained nearby to assist if needed. In the second, they wanted the robot to manage the entire activity independently for the child. However, our current design of the \textit{robot takeover mode} remains too ``parent-centered,'' hindering young children's independent use due to their limited reading skill, touchscreen proficiency, and attention spans. Many parents noted that older children might handle the system independently, suggesting that the \textit{robot takeover mode} should incorporate more interaction modalities, adjustable information density, and varying activity lengths to accommodate development needs of children as they mature. From an educational standpoint, younger children benefit from tasks that are neither too challenging nor too easy, as overly difficult tasks can lead to frustration and loss of confidence. Thus, when the \textit{robot takeover mode} is active, ensuring that the learning content presents an appropriate level of challenge to maintain engagement and support children's learning is essential.


\subsection{Design Implications} 
We proposed five design implications based on the discussion of the findings to inform future design for AI-assisted robot.

\subsubsection{Design Implication \#1: } \textit{Tailoring the Quality of LLM Models with Minimal Parental Feedback}.  
When parents have limited time or energy, they often provide minimal content review and may avoid using the system if they lack trust in the model or cannot supervise. This suggests that LLM-generated content must be trustworthy enough for parents to use with little oversight. Beyond employing advanced LLMs, designers should incorporate features that encourage parents to supervise the content considering their availability, \textit{e.g.,} summarization versus detailed view of all content. Additionally, integrating evaluations for factors like age appropriateness can increase parents' trust and reduce the likelihood of disuse.

\subsubsection{Design Implication \#2:} \textit{Explicit Design for Varying Levels of Control over LLM-Generated Content}.  
Parents engage with LLM-generated content differently depending on their circumstances—ranging from using it as is, reusing previously reviewed content, quickly skimming, or extensively customizing. This variation calls for a more explicit design of control options within the system. Instead of presenting all features (\textit{e.g.,} regenerate, edit) simultaneously, the interface should adapt to users' needs. Those skimming for a quick review could access a concise summary, while parents interested in customization might see detailed content and easy-to-use editing and regeneration tools.

\subsubsection{Design Implication \#3: } \textit{Mode-Switching as a Shift in User Center, Not Just Task Delegation}.
Currently, the mode-switching feature focuses on dividing tasks between the parent and the robot, but it should also reflect changing interaction dynamics between parent and child. For example, when switching from parent-led to parent-takeover mode, a parent may want to access LLM-generated content without the robot's presence. Yet, the current design presents content uniformly across modes. Similarly, shifting from robot-led to robot-takeover mode implies independent child use, but the system currently remains parent-centered. To address this, mode-switching should not only redistribute tasks but also adjust content presentation and interaction methods. In parent-takeover mode, content could be simplified for easier navigation, and in robot-takeover mode, interaction could shift from touch-based to voice-based, fostering a child-friendly environment without parental intervention.

\subsubsection{Design Implication \#4: } \textit{Maximizing Parent-Child Interaction Value through Appropriate Difficulty Levels}.
Mode-switching should also consider optimizing educational impact for different interaction dynamics by adjusting content difficulty. Parents generally excel at reading, explaining, and encouraging their children, while the robot's role is to supplement limited time and energy and offer alternative ways to engage children. When parents are involved, slightly more challenging content can increase the educational value of their engagement. Conversely, in child-robot interactions, the system should present content that children can handle independently, avoiding frustration or loss of confidence.

\subsubsection{Design Implication \#5: } \textit{Integrating Parenting Education into the System}.
Parents view the system not only as a tool for engaging their children, but also for improving their own parenting skills, \textit{e.g.,} effective questioning technique and pedagogical strategies. While familiar with preschool-level concepts, they may struggle to generate questions spontaneously or accurately gauge their child's abilities. To address this, the system should incorporate parenting education features. It could, for example, guide parents in developing more intuitive questioning techniques and provide tools to track and benchmark their child's progress. By doing so, the system empowers parents to focus their efforts and better support their children's learning.

\section{Limitations \& Future Work}\label{sec-7.4}

A notable limitation of our study is the lack of sociodemographic diversity in our participant sample. Specifically, (1) \textit{location:} all participants were based in the U.S. Midwest; (2) \textit{gender:} the 20 parents who participated in the study included only three fathers; (3) \textit{income:} the majority of families reported middle to high household incomes; and (4) \textit{education:} most parents had a Master's degree or higher. These factors may limit the broader applicability of our findings. Consistent with previous education studies, a predominance of mothers were included due to self-selection bias \cite{schoppe2013comparisons, mcbride1993comparison}, and we faced difficulties recruiting lower-income families \cite{nicholson2011recruitment}. Children from higher-income and more highly educated families tend to perform better academically \cite{sirin2005socioeconomic}. These may have influenced the types of support families wanted from the robot, which need to be addressed in future research. we aim to collaborate with local community centers, libraries and schools to reach a more diverse population in our future work.

In addition, while cost remains a common accessibility limitation, some educational robots, such as Miko,\footnote{Miko Robot: \url{https://shorturl.at/XSyM5}} are now priced similarly to smartphones, improving feasibility for families. Future work should design novel interaction paradigms for AI-assisted educational robots in public spaces (\textit{e.g.,} libraries, schools, museums) to broaden accessibility. Moreover, our brief home visits captured immediate reactions rather than long-term changes in parent-child interactions, trust in AI/robots, or learning outcomes. Extended research are needed to observe how these factors evolve as AI-assisted robots integrate into daily life.

Additionally, we acknowledge the limitation of simplifying parental involvement factors into binary levels, resulting in eight scenarios. This decision was practical, as more granular factors (\textit{e.g.,} three levels each; 27 scenarios) would yield in a large number of scenarios not manageable for human participants. Future work should explore methods for representing a continuous spectrum of factors to capture more accurate scenarios. Furthermore, our system currently targets literacy and math. Beyond these academic subjects, other domains like social-emotional skills and creativity warrant exploration. Investigating parent-AI-robot collaboration in these areas could yield broader insights. Finally, ethical concerns motivated our work (Section~\ref{sec-rw-2.2}), but were not our main focus. We presented some ethical issues in Section~\ref{sec-result-2} and discussed in Section~\ref{sec-dis-2}, but future research should delve more deeply into these topics. A comprehensive understanding of ethical considerations, from data sourcing to model training, will help ensure that AI-generated content meets appropriate standards for children.
\section{Conclusion}

In this paper, we conducted an in-home study with 20 parent-child pairs, each with a child aged 3--5 years old, to explore parental involvement scenarios, their use of LLM-generated learning content, and collaboration patterns with an educational robot in varying contexts. We developed a card-based activity kit, \texttt{SET}, to systematically map out parental involvement scenarios. Additionally, we introduced \texttt{PAiREd}, an LLM-assisted interface integrated with an educational robot, enabling parents to review and revise LLM-generated learning content while adjusting their level of involvement by delegating roles between themselves and the robot. Our findings revealed that parents seek additional guidance in pedagogical strategies, especially for advanced learning concepts. We also observed mixed attitudes toward AI-generated content, with parents prioritizing factors such as difficulty, variety, and quality during content review. Furthermore, we identified that parents vary in how they use AI-generated content, choosing to edit, skim, reuse, use directly, or even avoid it. They also appreciate the pedagogical insights the system offers throughout the process. Lastly, we uncovered preferences for collaborating with the robot across different modes, emphasizing design opportunities that go beyond task distribution, focusing single or multiple user perspectives, and adjusting content difficulty for each mode to maximize the value of interaction. Our work provides a novel approach to capturing parental involvement scenarios, offers deeper insights into parent interactions with AI-generated learning content, and presents an innovative interaction paradigm for dynamically adjusting involvement based on scenario changes.

\begin{acks}
This work was supported by the National Science Foundation awards 2202803. We thank Nathan White for his guidance with the technical development of our robot system. We also thank Amy Koike for her help in creating Figures 4 and 5.
\end{acks}

\balance
\bibliographystyle{ACM-Reference-Format}
\bibliography{Bibliography}


\end{document}